# Khmer Word Search: Challenges, Solutions, and Semantic-Aware Search

**Rina Buoy**     **Nguonly Taing**     **Sovisal Chenda**
Techo Startup Center (TSC)
{rina.buoy,nguonly.taing,sovisal.chenda}@techostartup.center

**Abstract**

Search is one of the key functionalities in digital platforms and applications such as an electronic dictionary, a search engine, and an e-commerce platform. While the search function in some languages is trivial, Khmer word search is challenging given its complex writing system. Multiple orders of characters and different spelling realizations of words impose a constraint on Khmer word search functionality. Additionally, spelling mistakes are common since robust spellcheckers are not commonly available across the input device platforms. These challenges hinder the use of Khmer language in search-embedded applications. Moreover, due to the absence of WordNet-like lexical databases for Khmer language, it is impossible to establish semantic relation between words, enabling semantic search. In this paper, we propose a set of robust solutions to the above challenges associated with Khmer word search. The proposed solutions include character order normalization, grapheme and phoneme-based spellcheckers, and Khmer word semantic model. The semantic model is based on the word embedding model that is trained on a 30-million-word corpus and is used to capture the semantic similarities between words.

## 1 Introduction

While the digital contents written in Khmer script have been growing rapidly over digital platforms, retrieving relevant contents is, otherwise, challenging [1]. Although major search engines such as Google support Khmer language, the integration has not taken into full account of Khmer script's features [1; 2].

The study by [2] shows Google search hits for the word ស្ត្រី vary from 23 thousands to 5 millions depending on the underlying character sequences. This issue is known as identical rendering.

As many digital platforms and applications such as e-commerce, e-learning, e-banking, delivery, and social media allow users to view content, provide user inputs, and search using Khmer script, there are, however, peculiar challenges associated with the Khmer language, which hinder system performance, usability, and user experience. The challenges addressed in this work are:

- Stacked vs unstacked syllable: If a user types បម្លែង (to convert) as a search query and the database contains only បំលែង , the search is unable to find any match since បម្លែង and បំលែង are treated as different words despite the fact that they are different spelling realizations of the same word.

- Spelling mistake: Most spelling mistakes are primarily due to typing errors. In some cases, a user may miss one or more vowels, or diacritics. In other cases, a user may mistakenly write ◌ី or ឡ្យ instead of ◌ិ or ល , respectively. For examples, ប្រសិទ្ធភាព(efficiency) and ប្រសិទ្ធីភាព , សាលារៀន (school) and សាឡ្យារៀន .

- Wrong part of speech: Ex. អភិវឌ្ឍ (v. - to develop) and អភិវឌ្ឍន៍ (n.).

- Identical rendering: ស្ត្រី (women) can be written in different sequences of characters. ស្ត្រី= ស+ប្រ+្ត+◌ី or ស+្ត+ប្រ+◌ី or ស+ប្រ+◌ី+្ត . This is problematic for string matching during the search.

- No existing model to find semantically-related words: In some cases, a user

knows what he/she is searching for, but may not know the exact keywords. In this case, being able to find semantically-related keywords is useful. For example, សំលៀកបំពាក់ (clothes) is semantically related to ខោ (pants), អាវ (shirt), ខោអាវ (dress), អាវធំ (blazer), ស្រោមជើង (socks) and the like.

We propose the following solutions to the above-stated challenges:

- Character sequence normalization: a user search query is normalized by reordering the characters in a specific order.

- Spellchecking: A user search query is checked for spelling mistake by using grapheme-based and/or phoneme-based spellcheckers. Grapheme-based and phoneme-based spellcheckers are able to suggest the possible corrections within a pre-defined edit distance. Phoneme-based spellchecker can be also used to identify different spell realizations of the same words as in the case of បម្លែង and បំលែង.

- Semantic modelling of words: a Khmer word embedding model was trained using machine learning algorithm to be able to locate semantically-related words. The model was trained on 1-million sentences corpus which has approximately 30 million words.

The following sections of the paper are organized as follows: In section 2, background of Khmer language and related work are provided. In section 3, we elaborate the detailed description of the proposed solutions. Section 4 explains models setup and training while section 5 and 6 provide results and discussion, followed by a conclusion in section 7.

## 2 Background

### 2.1 Khmer Script

Khmer (KHM) is the official language of the Kingdom of Cambodia. The Khmer script is used in the writing system of Khmer and other minority languages such Kuay, Tampuan, Jarai Krung, Brao and Kravet. Khmer language and writing system were hugely influenced by Pali and Sanskrit in early history [3; 4].

The Khmer script is an abugida that was descended from the Brahmic script. Unlike Latin-based languages, up to 2 consonants can be stacked below a base consonant using the alternate form (aka Coeng - foot) to form a consonant cluster [3; 4; 5]. ស្ត្រី (women), for example, is a single-syllable word that is formed a consonant cluster of 3 consonants (ស+្ត+្រ). Various diacritical signs can be positioned above a consonant for various spelling conventions. Toandakhiat (់) is used to silence a character as in the word - រាមកិរ្តិ៍ (ramayanak) [4].

Khmer script has a large inventory of dependent vowels that cannot stay alone by themselves and have to be attached to a base consonant. Orthographically, a dependent vowel can be placed to the left, right, above, below or around a base consonant [6; 4; 7]. Another group of vowels is called independent vowels and behaves like a consonant. However, independent vowels are hardly used in the modern Khmer writing system [4]. For example, ឮ (to hear) is an independent vowel as well as a valid word.

### 2.2 Series and Vowel Quality

Khmer consonants are grouped into two groups or series: 1st (a) and 2nd (o). Most consonants in one series have their counterpart in another series; otherwise, triisap and muusikatoan are used to convert from 1st-series to 2nd-series and vice versa. Series is one of the special features of Khmer script as it determines the vowel quality [6; 3; 4]. When ា is attached to ក, the resulting syllable is pronounced as /kɑɑ/ and when ា is attached to គ, the syllable is pronounced as /kie/. Nevertheless, vowels such ិ, េ, and ើ do not change their quality regardless of the attached consonant's series.

Some loanwords from Pali or Sanskrit are often pronounced differently from their orthographic representation [4]. សតិ (/saʔ.teʔ/ - consciousness), សំស្ក្រឹត (/saŋ.skrət/- Sanskrit).

### 2.3 Stacked vs Unstacked Syllable

According to Huffman, Khmer syllables can grouped into three categories namely: mono-syllables, disyllables and polysyllables [4; 3].

- Monosyllables has a format of C1C2C3VF. The onset can be made up of a cluster with up to three consonants. F is a final consonant which is optional. For example, ស្រា (alcohol), ក្ដាម (crab), ខ្មែរ (Khmer).

- A disyllabic word is formed by joining a minor syllable and a major syllable. The minor syllable is susceptible to extreme syllable reduction. For example, ចម្ការ (plantation), ចម្រៀង (song), កម្សាន្ត (leisure)

- Polysyllables is formed by combining multiple monosyllables. Polysyllables words are commonly loanwords.

Disyllabic words can have two spelling forms, namely, stacked (ព្យញ្ជនតម្រួត) or unstacked (ព្យញ្ជនរាយ) [3]. Both spelling forms can produce the same pronunciation but are not officially equivalent. The official spelling form of those wells is based on Chuon Nath dictionary. In practice, most people use both spelling forms interchangeably.

## 2.4 Identical Rendering

One of the major issues in Khmer Unicode is identical rendering [2; 4; 8]. Identical rendering refers to cases in which different sequences of the same characters are rendered as the same words. Taking the word ស្ត្រី as an example, it can be written in three different sequences of characters as shown above.

Identical rendering does not only affect search performance using Khmer script, but also introduces vulnerabilities of spoofing attacks [2; 4; 8]. Other sources of identical rendering are uses of consonant shifter, vowels of two Unicode points and identical rendering of subscript ត and ដ.

## 2.5 Khmer Spellchecker

Since a word can be represented in many different orthographic forms as discussed earlier, avoiding spelling mistake is difficult if not impossible even for experienced writers in the absence of a robust spellchecker. According to [9; 10], spelling mistakes can be grouped into 4 major categories:

- Using wrong key typing;
- Errors in pronunciation and spellings;
- Errors due to homonyms; and
- Using wrong words.

Most Khmer spellcheckers such as SSBIC [1] are grapheme-based and based on the concept of noisy channel model. Only candidate words within a pre-defined threshold distance are used as candidate corrections [11]. A large threshold results in many irrelevant corrections while a small one may not generate enough corrections. For example, រមកេ (/riem.kee/ - wrong) and រមកិរ្តិ៍ (/riem.kee/ - correct) has an orthographic edit distance of 6.

## 2.6 Khmer Word Semantic Models

### 2.6.1 Khmer WordNet

One of the useful features of search is an ability to search by meaning. This is known as semantic search. It is possible by using word semantic models such as WordNet. WordNet contains semantic relations and meanings of each word. However, there is, to the best of the authors' knowledge, no publicly available Khmer WordNet that can be used to establish word semantic relations. There is an ongoing effort by [12] to construct Khmer WordNet by translating Princeton English WordNet's synsets. However, the coverage is just about 31% of Khmer words.

### 2.6.2 Khmer Word Embedding Model

A word embedding is a learned representation capable of capturing context of a word, semantic and syntactic similarities and relations with other words. There exists a pre-trained word embedding model for Khmer language. The model is a FastText[2] model that was trained on Khmer Wikipedia corpus by using ICU tokenizer as a word segmenter. ICU tokenizer is a dictionary-based approach to identify words. Such approach is sensitive to spelling mistakes, unable to handle of out-of-vocabulary words, and does not take into account context [13; 14]. Khmer is highly analytic in morphology, and therefore, context is important in identifying token or word boundaries [15].

---

[1]https://github.com/sbbic/sbbic-khmer-spelling-checker-for-ms-word

[2]https://fasttext.cc/docs/en/crawl-vectors.html

## 3 Solutions to Khmer Word Search

### 3.1 Character Sequence Normalization

A Khmer word is made of a series of character clusters (KCC) or orthographic syllables [16; 13; 5; 7]. For example, សាលា (school) is made of two clusters namely: សា and លា. A KCC is made of a base consonant, one or more consonant subscripts, an dependent vowel, and a diacritical sign. The idea of KCC is that subscript consonants, dependent vowels, and diacritical signs must be preceded by a base consonant. A complete rule of construction a KCC is explained [10].

A KCC is normalized in the following steps:

1. Decomposition: KCC = [base consonant] + [consonant subscript(s)] + [diacritical sign] + [dependent vowel]

2. Sorting: sort consonant subscripts if there are more than one.

3. Recombination: [base consonant] + [consonant subscript(s)] + [diacritical sign] + [dependent vowel] = KCC

By applying the normalization steps on different sequences of the word ស្ត្រី, the resulting normalized sequence is ស + ្ត + ្រ + ី.

The limitation of the normalization steps is its inability to handle identical rendering caused by misuse of consonant subscript ្ត and ្ដ. The re-ordering of consonant subscripts may not necessarily follow linguistic rules.

### 3.2 Khmer Spellcheckers - from Grapheme to Phoneme

We implemented a Khmer grapheme-based spellchecker by using SymSpell [3] that relies only on deletion operation to generate candidate corrections as illustrated in Figure 1.

---
[3]https://github.com/wolfgarbe/SymSpell

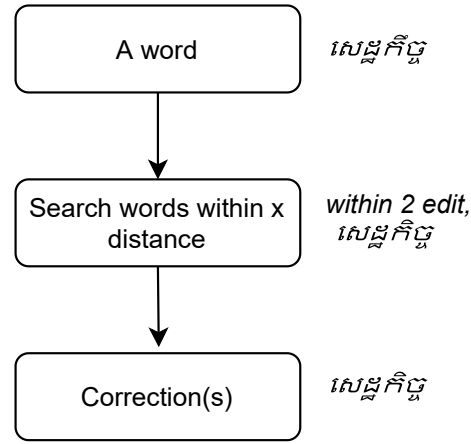

Figure 1. Grapheme-based spellchecker

The grapheme-based spellchecker works reasonably well when correct word(s) are within a close edit distance from an input misspelled. However, in Khmer language, discrepancies between spelling and pronunciation are not uncommon due to Pali/Sanskrit loanwords, the use of Toandakhiat, and the interchangeability of stacked and unstacked syllable writing. Thus, the correct word(s) can be far from an input misspelled word orthographically, but close phonemically. A phoneme-based spellchecker is, hence, proposed to complement the grapheme-based counterpart.

A phoneme-based spellchecker requires a grapheme-to-phoneme (G2P) model to convert grapheme to phoneme. The candidate corrections generation process is done by using SymSpell in phonemic space before converting back to grapheme space. The process flow is given in Figure 2.

We implemented a deep learning-based Khmer G2P model using an encoder-decoder (also known as Seq2Seq) network on the publicly available dataset[4] released by Google. A Seq2Seq model is suitable for tasks in which both output and input sequence length are variable [17]. This is the case of G2P task. The trained encoder-decoder model achieved a character error rate (CER) of 6.5% vs 7.5% by using a weighted finite-state transducer (WFST) that was used in [18]. An example of Khmer words with multiple spelling forms along with their corresponding phonemes is

---
[4]https://github.com/google/language-resources

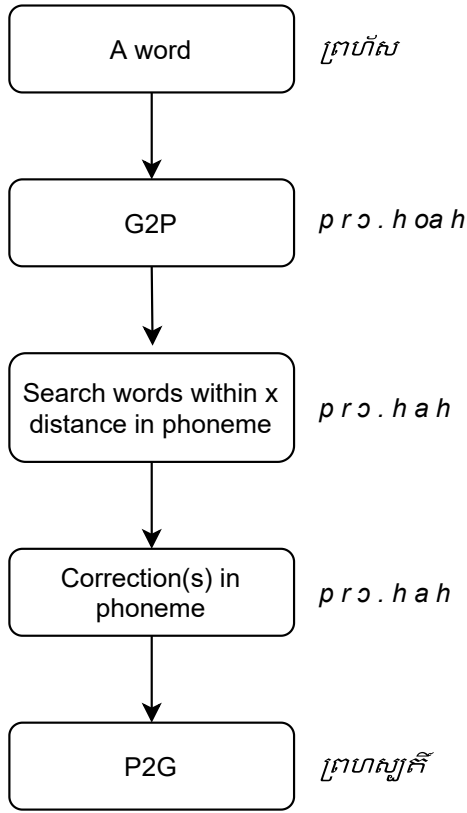

Figure 2. Phoneme-based spellchecker

given below:

- ជ័យជំនះ (victory) - /c e j . c u m . n ea h/
- ជ័យជម្នះ (victory) - /c e j . c u m . n ea h/

The performance comparisons of different spellcheckers on a few selected misspelled words are given in Figure 1, suggesting that both of our spellcheckers outperform SBBIC even in challenging cases.

## 3.3 Semantic Modelling of Khmer Words

FastText uses a sub-word model by representing each word as itself plus a combination of sub-words [19]. FastText can, therefore, compute representations for previously unseen words or words with spelling errors.
.

### 3.3.1 Data Collection Embedding Model Training

The primary data sources are Wikipedia texts and online news articles from various sites. The aggregated dataset is made up of approximately 1 million sentences that is about 30 million words.

The embedding model was trained by using FastText that is a library for efficient learning of word representations and sentence classification. The model was trained on Google Colab's virtual machine that is Intel(R) Xeon(R) CPU @ 2.30GHz with a ram of 32GB. It took about 12 minutes using FastText's default settings. The result of FastText binary file is approximately 800MB.

### 3.3.2 Locating Neighboring Words

One of the common similarity metrics between two words is cosine similarity [11]. Armed with the trained FastText embedding and cosine similarity, it is possible to locate the nearest neighbor words for a given input word. A few selected words along with their nearest neighbor words are given in Table 2.

Table 2 shows that the trained embedding model manages to capture the semantically similar and topically related words. The results in Table 3 from the pre-trained model using ICU tokenizer show some mix-ups of undesirable characters, which indicate lack of text pre-processing and cleaning, and it is unable to find the neighbor words for the word កូវីដ (Covid-19) which is a relatively recent loanword.

One of the interesting features is that both embedding models return spell errors or equivalent spelling forms as neighbor words as in the case of សម្លៀកបំពាក់ (clothes). This illustrates that a subword embedding model can be used to locate spelling errors.

## 4 Experiment with Khmer Information Retrieval (IR) System

We applied the above proposed solutions to an experimental Khmer information retrieval system. We set up an Elastic-search database server containing around 10000 news articles. Each article was segmented with the in-house word segmenter before applying sequence normalization and finally indexing.

The idea of this experiment is to prove that by applying the proposed solutions to expand an input query or word, the search engine can generate more relevant search results than it can otherwise generate with the raw input query. If the input query is misspelled, spellcheckers can check, correct and return correction(s) along with other equivalent spelling forms if there is any. For example, if the input query is ចាំរៀង (a misspell of ចំរៀង - song), spellcheckers return ចម្រៀង and ចំរៀង. If both ចម្រៀង and ចំរៀង do not exist in the database, the input query can be further expanded by semantically-related words.

| Missplelled words | SBBIC Spellchecker | Grapheme Spellchecker | Phoneme Spellchecker |
|---|---|---|---|
| សេដ្ឋកិច្ច | សេដ្ឋកិច្ច | សេដ្ឋកិច្ច | សេដ្ឋកិច្ច |
| ព្រហ៊ិស | - | - | ព្រហស្បត៍ |
| ចាំរៀង | ចំរៀង | ចំរៀង | ចម្រៀង, ចំរៀង |
| កាម់ចាត់ | កាត់តាម | កម្ចាត់ | កម្ចាត់, កំចាត់ |
| សម់លាញ | សម្លាញ់ | សម្លាញ់ | សម្លាញ់, សំឡាញ់ |

Table 1. Comparisons of different spellcheckers for selected Khmer words

| Words | Nearest Neighbors |
|---|---|
| ក្វីត | វ័រស,ក្វរីដ,ក្វរ៌ណា,ក្វរី,ក្វរីខ,ក្វរ៌ណាវ័រស,ក្វរ៌ណាមួយ,វាតឲ្យាត,ជំងឺក្វរីដ,ក្វរ៌ណា |
| ឡាន | រថយន្ត,ម៉ូតូកង់បួន,រថយន្ត,ម៉ូតូ,ម៉ូតូកង់,ការាំស,សោរថយន្ត,សណ្ណោង,អាតុប,តាក់ស៊ី |
| សម្លៀកបំពាក់ | សំលៀកបំពាក់,សម្លៀក,សម្លៀបំពាក់,ស្លៀកពេល,ការស្លៀក,ស្លៀកជើង,ខោអាវ,ម៉ូត,រូប,ការស្លៀកពាក់ |
| ធុរកិច្ច | ធុរៈកិច្ច,អាជីវកម្ម,វានុវត្តភាព,ធុរ,កសិពាណិជ្ជកម្ម,វានុវត្តន៍,ភាពប្រកួតប្រជែង,កលានុវត្តភាព, កាទ្បានុវត្តភាព,វានុវត្ត |
| សៀមរាប | សៀមរាម,ខេត្ត,បាត់ដំបង,បាត់ដំបង,បន្ទាយមានជ័យ,កំពង់ធំ,អង្គរ,តាកែវ,កំពង់ឆ្នាំង,ខេត្ត |

Table 2. Nearest neighbors located by our embedding model for selected Khmer words

| Words | Nearest Neighbors |
|---|---|
| ក្វីត | N/A |
| ឡាន | ស្យូមុន,TMnr,Catalan's,អេដហ្គារ,ជិះឡាន,Marulić,បើកឡាន,តែកុង,Daihatsu,BMW |
| សម្លៀកបំពាក់ | សម្លៀកបំពាក់,សម្លៀក,សម្លៀ,សំលៀកបំពាក់,របូតបាញ,ស្លៀកពាក់,ខោអាវ,ខសំលៀកបំពាក់,អាវធំ,រូប |
| ធុរកិច្ច | ធុរកិច្ចNo,សណ្ឋារកិច្ច,ធុរជន,ការៈកិច្ច,បជំសណ្ឋាកិច្ច,សហត្រិនភាព,ការកិច្ច,អ្នកសេដ្ឋកិច្ច, រដ្ឋOntarioនៃ,អាជីវកម្ម |
| សៀមរាប | div_5,am0សៀមរាប,ខេត្តសៀមរាប,pm0សៀមរាប,បាត់ដំបង,ខេត្ត,គោកចក,ចុងយ៉ោស, មាន១០ឃុំ,សៀមរាម |

Table 3. Nearest neighbors located by the pre-trained embedding model for the same words

## 5 Results

Elastic-search uses a Boolean model to locate matching documents for a given query and then, a scoring function that is based on TF-IDF concept (term frequency-inverse document frequency) for ranking relevancy. We measured the number of search hits when a query is not expanded and when it is expanded by applying the proposed solutions. The following scenarios were being evaluated:

1. Character sequence normalization: Table 4 shows the search hits for the cases before and after applying normalization. By applying normalization, it does not matter the character sequences of which a word ស្ត្រី is spelled, search engine can return maximum search hits.

2. Expanding a query with spellcheckers: In this case, a query is a misspelled word. Spellcheckers (both grapheme-based and phoneme-based checkers) are used to check and provide correction(s) along alternative spelling forms if there is any. Then, the query is expanded by correction(s) and alternative forms. Table 5 shows that with a misspelled query, the search engine is unable to find any matching documents. It also suggests that in case of multiple alternative forms exists as in the case of ចំរៀង vs. ចម្រៀង, សំលៀកបំពាក់ vs. សម្លៀកបំពាក់, concatenating them via OR operator returns maximum search hits. Table 5 gives total search hits of the individual queries (ចំរៀង, ចម្រៀង) , the sum of which is not necessarily equal to that of the concatenated query (ចំរៀង OR ចម្រៀង). This is due to the fact that within a single document, both ចំរៀង and ចម្រៀង exist. This indicates common spelling interchange-ability of some Khmer words.

| Queries | Search Hits |
|---|---|
| ស្ត្រី (ស+្រ+្ត+ី) | 0 |
| ស្ត្រី (ស+្ត+្រ+ី) | 252 |
| ស្ត្រី (ស+ី+្ត+្រ) | 0 |
| ស្ត្រី after normalization | 252 |

Table 4. Search hits before vs. after normalization

| Queries | Search Hits |
|---|---|
| ចាំរៀង | 0 |
| ចំរៀង | 14 |
| ចម្រៀង | 231 |
| ចំរៀង OR ចម្រៀង | 239 |
| សំលាក់បំពាក់ | 0 |
| សំលៀកបំពាក់ | 59 |
| សម្លៀកបំពាក់ | 58 |
| សំលៀកបំពាក់ OR សម្លៀកបំពាក់ | 111 |

Table 5. Search hits before vs. after spellchecking

3. Expanding a query with semantically-related words: An input query is expanded by including words to which the query is semantically related. Let's use a word, សម្លៀកបំពាក់ (clothes) as example. Semantically-related words include ស្បែកជើង(shoes) ខោអាវ(clothes) ម៉ូត(style) រូប(robe) and ការស្លៀកពាក់(putting on clothes). The increasing numbers of search hits for incrementally-expanded queries are given in Table 6.

## 6 Discussion

The experimental results suggest that the proposed solutions can mitigate peculiar problems associated with Khmer word search by applying character sequence normalization and spell checking. A trained Khmer word embedding model makes search become semantic-aware. Therefore, search engine can find more matching documents for a given input query than it otherwise can with the raw query. However, there remain additional problems worth further attention and research. These problems are discussed below.

| Queries | Search Hits |
|---|---|
| សំឡេកបំពាក់ OR សម្លេកបំពាក់ | 111 |
| សំឡេកបំពាក់ OR សម្លេកបំពាក់ OR ស្បែកជើង | 149 |
| សំឡេកបំពាក់ OR សម្លេកបំពាក់ OR ស្បែកជើង OR ខោអាវ | 179 |
| សំឡេកបំពាក់ OR សម្លេកបំពាក់ OR ស្បែកជើង OR មួក OR ខោអាវ | 210 |
| សំឡេកបំពាក់ OR សម្លេកបំពាក់ OR ស្បែកជើង OR មួក OR ខោអាវ OR រូប | 218 |
| សំឡេកបំពាក់ OR សម្លេកបំពាក់ OR ស្បែកជើង OR មួក OR ខោអាវ OR រូប OR ការស្លៀកពាក់ | 226 |

Table 6. Search hits for incrementally-expanded queries

### 6.1 Identical Rendering

The proposed character sequence normalization solves an identical rendering issue caused by permutation of consonant subscripts. Identical rendering can also be caused by the following cases [8; 4]:

- Consonant shifter (aka series or register shifter);
- Two Code-Points Vowels; and
- Consonant subscript ត and ដ.

### 6.2 Inconsistent Spelling and Pronunciation

Pronunciations of some Khmer words do not necessarily follow its orthographic representation. Here are some cases leading to the inconsistencies between spelling and pronunciation:

- Loanwords from Pali and Sanskit: សាស្ត្រ (science) is pronounced as /sah/ while សាស្ត្រា (palm leaf manuscript) is pronounced as /sah.straa/.

- Ambiguities between orthographic and morphological syllable: For example, លើក ( /ləə.kɔɔ/ - lift) can be a single syllable word which is composed of two orthographic syllables - លើ (above) and ក (neck).

- Ambiguities between consonants ត and ដ: Some words are written with consonant ត but are pronounced with consonant ដ. For example, តម្លៃ is pronounced as /ɗɑm.laj/ instead of /tɑm.taj/.

These inconsistencies may hinder the generalization of G2P model.

### 6.3 Spelling Errors in Embedding Model

As word segmentation is required to prepare dataset for training a Khmer word embedding model, errors due to segmentation or actual spelling errors are manifested in the resulting embedding model. The embedding model, thus, returns neighboring words that are spelling errors for a given input query. This behavior can be mitigated by performing post spellchecking on the returned neighboring words.

Word segmentation and spell correction are tightly coupled processes. Spelling errors lead to wrong segmentation and vice versa.

## 7 Conclusion

In this paper, we present the peculiar challenges associated with search using Khmer script, that are caused by identical rendering, spelling errors and multiple spelling forms. We, thus, proposed characters sequence normalization and robust Khmer spellcheckers that use both grapheme and phoneme. To perform spellchecking phonemically, we trained a Khmer G2P model on the publicly available Google dataset. To make search semantic-aware, a Khmer word embedding model was trained on the dataset of around 1 million sentences ( 30 millions words). The resulting embedding model can locate semantically related words by applying cosine similarity on word vectors. We apply the proposed solutions on an experimental information retrieval using an Elasticsearch database server. The experiment results show that search engine can find more matching documents with expanded queries generated by the solutions than it otherwise can with a raw query.